%% file: sample-sigconf.tex
\pdfoutput=1

\documentclass[sigconf]{acmart}

\usepackage{booktabs} 
\usepackage{multirow}
\usepackage{graphicx} 
\usepackage{float} 
\usepackage{subfigure} 
\usepackage{algorithm}
\usepackage{algorithmic}

\renewcommand\footnotetextcopyrightpermission[1]{} 
\begin{document}
\title{Multi-View Non-negative Matrix Factorization Discriminant Learning via Cross Entropy Loss}
\author[mymainaddress]{Jian-wei Liu, Yuan-fang Wang, Run-kun Lu, Xionglin Luo}

\affiliation{Department of Automation, College of Information Science and Engineering,
China University of Petroleum , Beijing, Beijing, China}

\begin{abstract}
Multi-view learning accomplishes the task objectives of classification by leverag-ing the relationships between different views of the same object. Most existing methods usually focus on consistency and complementarity between multiple views. But not all of this information is useful for classification tasks. Instead, it is the specific discriminating information that plays an important role. Zhong Zhang et al. explore the discriminative and non-discriminative information exist-ing in common and view-specific parts among different views via joint non-negative matrix factorization. In this paper, we improve this algorithm on this ba-sis by using the cross entropy loss function to constrain the objective function better. At last, we implement better classification effect than original on the same data sets and show its superiority over many state-of-the-art algorithms.
\end{abstract}

%
%





\maketitle

\input{samplebody-conf}

\bibliographystyle{ACM-Reference-Format}
\bibliography{sigproc}

\end{document}

%% file: samplebody-conf.tex
\section{Introduction}

In many scientific data analysis tasks, data are often collected through different measuring methods, such as various feature extractors or different sensors, as usually the single particular measuring method cannot comprehensively extract all the information of the object or entities. In this case, the features of each example can be naturally divided into groups, each of which can be considered as a view. For instance, for images and videos, color features and texture features can be regarded as two views. It is important to make good use of the information from different views. Multi-view learning is a branch of machine learning which studies and utilizes the information and relationship between views.

There are also many entities represented with different views in the real world, such as web pages~\cite{1,2}, multi-lingual news~\cite{3,4,5}, and neuroimaging~\cite{6,7,8}. And most of relevant literatures of multi-view learning regard the consistency and complementarity as two main underlying properties of the multi-view data~\cite{9}, which like the bridges to link all views together~\cite{10,11,12}. Consistency assumptions indicate that all views share consistent information. Obviously, using only consistent information to take advantage of multi-view data is not enough, because each view also contains additional knowledge that other views do not have~\cite{1,13,14}. Therefore, the complementarity of views is another important properties of learning multi-view data, which is worth developing and utilizing.

However, it is a doubt that whether the consistent and complementary information really always support a better classification performance. Because some experiences of previous experiments show that the predictive performance of multi-view data may even be worse than only using single-view data in some real data sets. Zhang Zhong et al. hold point of  view  that the multi-view data contains parts that are helpful and unhelpful for classification, which are called discriminate and non-discriminate information and the consistent or complementary information does not learn discriminative information directly~\cite{15}. The classifier constructed by multi-view data may give an even worse classification performance if the learned consistent and complementary information contains no clear discriminative information.

In this paper, we propose multi-view non-negative matrix factorization discriminant learning via cross entropy loss. we distinguish discriminative and non-discriminative information existing in the consistent and complementary parts, and only use discriminative information for classification. More specifically, multi-view data is factorized into common part shared across views and view-specific parts existing within each view in the usual multi-view learning manner. Besides, for both common part and view-specific part, they are further factorized into discriminative part and non-discriminative part. In this situation, each view of data is factorized into four parts: the common discriminative part, the common non-discriminative part, the specific discriminative part and the specific non-discriminative part.

In previous research for multi-view learning, mean square error is habitually used as a loss function for predicting labels, but the truth is not the case. As the training results of gradient descent show, using mean squared error term to predict the output labels has a certain degree of defects. Its partial derivative value of mean squared error term is very small when the probability of output labels is close to 0 or 1, which could cause the value of partial derivative to almost vanish at the beginning of training. In order to speed up the training processes With the decrease of error between predicted label and actual label values, we choose the cross-entropy loss function instead of mean squared error. Cross-entropy loss function is usually used to measure the performance of a classification model whose input is a probability value between 0 and 1. And cross-entropy loss increases as the predicted probability diverging from the actual label. Moreover, a supervised constraint is added objective function to guide the joint NMF factorization to obtain better discriminative parts.

To find the optimal decomposition, we follow the block coordinate descent framework~\cite{16} to solve our proposed objective function. And only the derived discriminative parts from common part and view-specific parts are used to construct a classifier. Finally, experimental results on seven real-world data sets verify the effectiveness of our proposed method.

In conclusion, our contributions are summarized as follows:

(1)We propose a new multi-view non-negative matrix factorization discriminant learning algorithm, we dub Multi-View non-negative matrix factorization Discriminant Learning via Cross Entropy Loss(MV-DLCSL), which utilizes the discriminative information of data to classify, and outperforms many state-of-the-art algorithms on seven real-world data sets.

(2)To our best knowledge, we introduce cross-entropy loss into the non-negative matrix factorization framework for the first time, which is more reasonable objective function to measure predicting output label errors.

(3)To find the optimal decomposition, we follow the block coordinate descent framework to solve our proposed optimization problem. However, the cross-entropy loss function contains the nonlinear softmax function, which normalized the predicting output label probability. For better preforming the block coordinate descent, we devise approach to obtain the derivative of the cross-entropy loss function for each matrix vectors.

(4) In order to visually compare the difference between cross entropy loss and mean squared error, we use a visualization tool, t-SNT, which is very suitable for high-dimensional data dimensionality reduction to 2D or 3D for visualization. According to the result, our proposed method that use cross entropy as loss function is able to find the feature matrix more accurately than the mean squared error function.

\section{Related Work }
Two representative works of multi-view learning in the early days are canonical correlation analysis (CCA) (Hotelling 1936)~\cite{17} and co-training (Blum and Mitchell 1998)~\cite{1}. They represent two core ideas for dealing with multi-view problems through consistency and complementarity.

Studies in exploiting consistency generally looks for commonalities between multiple views, which have minimum disagreement. Canonical Correlation Analysis related algorithms~\cite{10,18,19,20,21} project two or more views into latent subspaces by maximizing the correlations among projected views. Spectral methods~\cite{5,22,23,24,25} use a weighted summation to merge graph Laplacian matrices from different views into one optimal map for further clustering or embedding. Matrix factorization based methods\cite{4,11,26} jointly factorize multi-view data into a common centroid representation by minimizing the overall reconstruction loss of different views. In addition, multiple kernel learning (MKL)~\cite{27} can also be viewed as taking advantage of the consistency of different views, where each view is mapped to the new space using kernel tricks, and then all kernel matrices are combined to unify the kernel by minimizing predefined objective functions.

Another method is to explicitly preserve complementary information of different views. The co-training algorithms~\cite{1,28,29,30} treat each view as complementary. Generally, it iteratively trains two classifiers on two different views, and each classifier generates its complementary information to help other classifiers train in the next iteration.

In summary, most existing multi-view learning algorithms mainly focus on learning consistency and complementarity from multi-view data. Zhong Zhang et al.~\cite{15} break away from convention and explores the discriminative and non-discriminative information existing in common and view-specific parts among different views via joint non-negative matrix factorization, which provides novel ideas for multi-view learning.

\section{The Proposed Method }

\subsection{Non-negative Matrix Factorization}

Given a non-negative matrix $X\in \mathbb{R}_{+}^{m\times n}$, which represents $n$ examples with   features. NMF aims to find non-negative matrix factors $W\in \mathbb{R}_{+}^{m\times k}$ and $H\in \mathbb{R}_{+}^{n\times k}$such that:
\begin{equation}
  X \approx  W{H^T}
\end{equation}
Then the objective function can be written as follows:
\begin{equation}
  \begin{aligned}
  \underset{W,H}{\mathop{\arg \min }}\,\left\| X-W{{H}^{T}} \right\|_{F}^{2}\\
  s.t.\quad W,H\ge 0
  \end{aligned}
\end{equation}
where $| \cdot |{|_F}$ denotes the Frobenius norm. Note that the original data matrix is a linear combination of all column vectors in $W$ with weight of corresponding column vec-tors in $H$. Therefore, $W$ and $H$ are usually called the basis matrix and the coefficient matrix respectively.
For multi-view data, the objective function learning representation of NMF-based approaches is as follows:
\begin{equation}
  \begin{aligned}
  \mathop {\arg \min }\limits_{W,H} \sum\limits_{v = 1}^{{n_v}} {\left\| {{X^{(v)}} - W{H^T}} \right\|_F^2 + \Phi (W,H)}\\
  s.t.\quad W,H \ge 0
  \end{aligned}
\end{equation}
where ${{n}_{v}}$ denotes the number of views, and ${{X}^{(v)}}$ denotes the data matrix of $v$-th view. $\Phi (\cdot )$ is a regularization term for $W$ and $H$.

\subsection{Multi-view Learning via DICS}
There are consistency and complementary among different multiple views. Therefore, DICS algorithm decomposes the multi-view data matrix into two parts: common part and view-specific parts. As other approaches~\cite{31,32,33,34}, define ${{W}_{C}}$ represents the common subspace shared by all views and $W_{S}^{(v)}$ represents the specific subspace for the $v$-st view. Therefore, the data matrix of each view can be written as ${{X}^{(v)}}={{W}_{C}}H_{C}^{T}+W_{S}^{(v)}H{{_{S}^{(v)}}^{T}}$. In order to learn the discriminate information, not only the data matrix is divided into the common and view-specific parts, but also each part of data matrix is divided into the discriminative and non-discriminative part as follows:
\begin{equation}
\begin{array}{l}
  W = [{W_{CD}}\,{W_{CN}}\,W_{SD}^{(v)}\,W_{SN}^{(v)}]\\
  H = [{H_{CD}}\,{H_{CN}}\,H_{SD}^{(v)}\,H_{SN}^{(v)}]
\end{array}
\end{equation}
where ${{W}_{CD}}$ and ${{H}_{CD}}$represent the common discriminate matrixes respectively, while${{W}_{CN}}$ and ${{H}_{CN}}$ separately represent the common non-discriminate matrixes. Similarly, $W_{SD}^{(v)}$ and $H_{SD}^{(v)}$ respectively indicate the view-specific discriminate parts, while $W_{SN}^{(v)}$ and $H_{SN}^{(v)}$ separately indicate the view-specific non-discriminate parts.
Afterwards, the discriminate matrixes is used to predict the output label through a linear projection matrix $B=[{{B}_{CD}}\,B_{SD}^{(v)}]$, which is corresponding to ${{H}_{CD}}$ and $H_{SD}^{(v)}$. Therefore, the objection function of DICS is further reformulated as follows:
\begin{equation}
  \begin{aligned}
  \mathop {\arg \min }\limits_{W,H,B} \sum\limits_{v = 1}^{{n_v}} {\left\| {{X^{(v)}} - W{H^T}} \right\|_F^2 + \Phi (W,H)}  \\
  + \gamma \left\| {Y - \left[ {{B_{CD}}\,B_{SD}^{(v)}} \right]\left[ {\begin{array}{*{20}{c}}
  {H_{CD}^T}{H_{SD}^{(v)T}}
  \end{array}} \right]} \right\|_F^2 \\
  s.t.\quad W,H \ge 0
\end{aligned}
\end{equation}
where $Y\in {{\mathbb{R}}^{c\times n}}$ is the label matrix, c is the number of classes and n is the number of data instances. Moreover, ${{y}_{i,j}}=1$if the $j$-st instance belong to class $i$and otherwise ${{y}_{i,j}}=0$.

\subsection{Improved Objective Function via Cross-Entropy Loss}
\begin{figure}[htbp]
\centering
\includegraphics[scale=0.7]{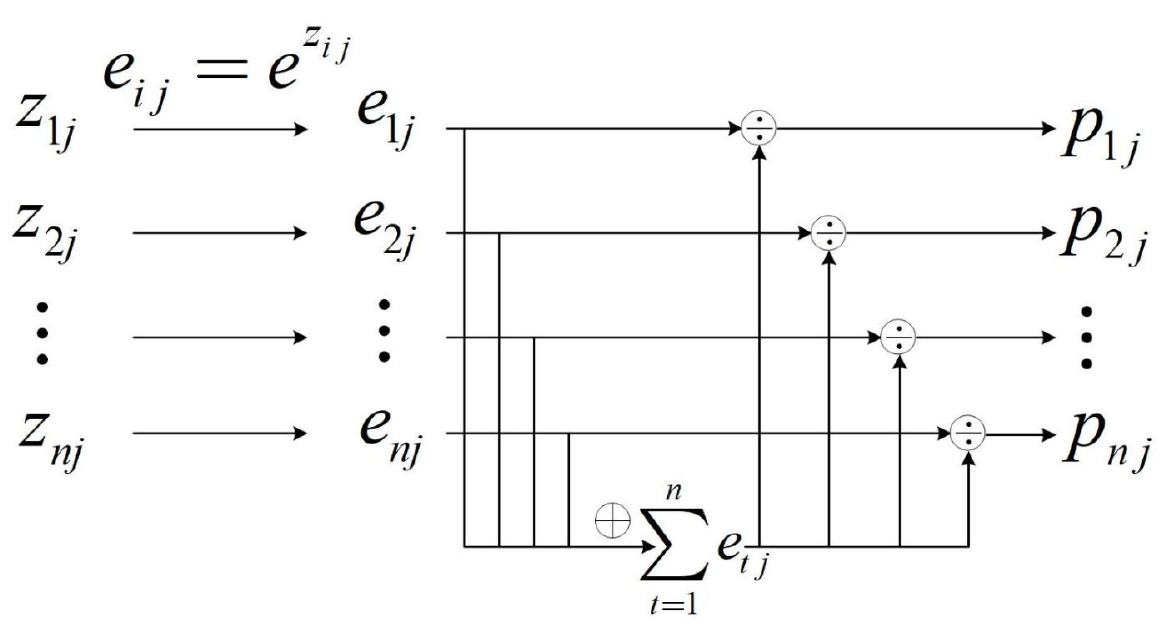}
\caption{Schematic diagram of the softmax function}
\label{Fig.softmax}
\end{figure}

However, as training results of gradient descent show in section 4 , the mean squared error term for predicting the output label has a certain degree of deficiencies. Its partial derivative value is very small when the output probability value is close to 0 or 1, which could suffer from the partial derivative value to almost disappear at the beginning of training. In order to improve the training convergence speed in time with decrease of the error between predicted and actual values, the cross-entropy loss function is selected instead of the mean squared error.
Cross-entropy loss function is usually used to measure the distance between two probability distributions whose input is a probability value between 0 and 1. And cross-entropy loss increases as the predicted probability diverging from the actual label.
For multi-classification tasks, the form of the cross-entropy loss function is as follows:
\begin{equation}
  J =  - \sum\limits_{i = 1}^c {{y_i}\log ({p_i})}
\end{equation}
Where $c$ denotes the number of classes, ${{y}_{i}}$ is an probability distribution of real labels, if the instance belong to the class i, ${{y}_{i}}$ is 1 otherwise 0. And ${{p}_{i}}$ is the predicted probability that the observed example belongs to class i.
Therefore, the improved objective function could be written as follows:
\begin{equation}
  \begin{aligned}
  \mathop {\arg \min }\limits_{W,H,B} \sum\limits_{v = 1}^{{n_v}} {\left\| {{X^{(v)}} - W{H^T}} \right\|_F^2 + \alpha {{\left\| {W_D^T{W_D}} \right\|}_{1,1}} + \beta {{\left\| {{H_D}} \right\|}_{1,1}}} \\
  - \gamma \sum\limits_{j = 1}^n {\sum\limits_{i = 1}^c {{y_{ij}}\ln {p_{ij}}} }
\end{aligned}
\end{equation}
Where $\alpha ,\beta ,\gamma $are non-negative parameters to balance the regularization term and ${{n}_{v}}$ denotes the number of views, $n$ denotes the number of instances, $c$ denotes the number of classes.
the ${{\left\| \,\cdot \, \right\|}_{1,1}}$ is a ${{L}_{1,1}}$ norm constraint on the discriminative matrix ${{W}_{D}}$, this second term in objective function can be factorized into two parts:
\begin{equation}
  {\left\| {W_D^T{W_D}} \right\|_{1,1}} = \sum\nolimits_i {w_{Di}^T{w_{Di}}}  + \sum\nolimits_{i \ne j} {w_{Di}^T{w_{Dj}}}
\end{equation}
the second term in objective function encourage basis vectors to be as orthogonal as possible, and which reduces the redundancy of discriminative bases. the third term in objective function is used to prevent overfitting, What's more, using ${{L}_{1,1}}$ norm constraint on ${{H}_{D}}$ makes the discriminate coefficients sparse. The reason is that data points of different classes should not possess identical basis vectors.
${{p}_{ij}}$ is a predictive label normalized by softmax function. Specifically, its form is as follows:
\begin{equation}
  {p_{ij}} = \frac{{{e^{\sum\nolimits_{k = 1}^{{k_1}} {{b_{C{D_{i,k}}}}{\rm{\cdot}}{h_{C{D_{j,k}}}}}  + \sum\nolimits_{k = 1}^{{k_3}} {b_{S{D_{i,k}}}^{(v)}{\rm{\cdot}}h_{_{S{D_{j,k}}}}^{(v)}} }}}}{{\sum\nolimits_{t = 1}^n {{e^{\sum\nolimits_{k = 1}^{{k_1}} {{b_{C{D_{t,k}}}}{\rm{\cdot}}{h_{C{D_{j,k}}}}}  + \sum\nolimits_{k = 1}^{{k_3}} {b_{S{D_{t,k}}}^{(v)}{\rm{\cdot}}h_{_{S{D_{j,k}}}}^{(v)}} }}} }}
\end{equation}
Thus the objective function can be written as follows in the form of element-wise and vectors:

\begin{equation}
  \begin{aligned}
  \begin{array}{l}
f(W,H,B) = \sum\limits_{v = 1}^{{n_v}} {{\rm{||}}{X^{(v)}} - \sum\limits_{i = 1}^{k1} {{w_{C{D_i}}}h_{C{D_i}}^T}  - \sum\limits_{i = 1}^{k2} {{w_{C{N_i}}}h_{C{N_i}}^T} } \\
\quad \quad \quad \quad  - \sum\limits_{i = 1}^{k3} {w_{S{D_i}}^{(v)}h_{S{D_i}}^{(v)T}}  - \sum\limits_{i = 1}^{k4} {w_{S{N_i}}^{(v)}h_{S{N_i}}^{(v)T}} {\rm{||}}_F^2\\
\quad \quad \quad \quad {\rm{ + }}\alpha (\sum\limits_{i = 1}^{k1} {\sum\limits_{j = 1}^{k1} {w_{C{D_i}}^T{w_{C{D_j}}}} }  + \sum\limits_{i = 1}^{k3} {\sum\limits_{j = 1}^{k3} {w_{S{D_i}}^{(v)T}w_{S{D_j}}^{(v)}} } \\
\quad \quad \quad \quad  + 2\sum\limits_{i = 1}^{k1} {\sum\limits_{j = 1}^{k3} {w_{C{D_i}}^Tw_{S{D_j}}^{(v)}} } )\\
\quad \quad \quad \quad  + \beta {1_{1 \times n}}\left( {\sum\limits_{i = 1}^{k1} {{h_{C{D_i}}}}  + \sum\limits_{i = 1}^{k3} {h_{S{D_i}}^{(v)}} } \right) - \gamma \sum\limits_{j = 1}^n {\sum\limits_{i = 1}^c {{y_{ij}}\ln {p_{ij}}} }
\end{array}
  \end{aligned}
\end{equation}
where the subscript $i$denotes the $i$-th row and the subscript $j$denotes the $j$-th column of the corresponding matrix. And ${{1}_{1\times n}}$ is a row vector of length $n$ with all elements 1.
By fixing all column vectors except the one we want to update, we can obtain the convex sub-problem respect to it, then solve it based on the block coordinate descent framework. In order to facilitate the derivation, the objective function is divided into two parts: the nonlinear part of cross-entropy loss function and the other linear parts.
The nonlinear part is written as follows:
\begin{equation}
\begin{array}{l}
L =  - \gamma \sum\limits_{j = 1}^n {\sum\limits_{i = 1}^c {{y_{ij}}\ln {p_{ij}}} } \\
{p_{ij}} = \frac{{{e^{{z_{ij}}}}}}{{\sum\nolimits_{t = 1}^n {{e^{{z_{tj}}}}} }}\\
{z_{ij}} = \sum\limits_{k = 1}^{{k_1}} {{b_{C{D_{i,k}}}}{\rm{\cdot}}{h_{C{D_{j,k}}}}}  + \sum\limits_{k = 1}^{{k_3}} {b_{S{D_{i,k}}}^{(v)}{\rm{\cdot}}h_{_{C{D_{j,k}}}}^{(v)}}
\end{array}
\end{equation}
Take the partial derivative of L to ${{h}_{C{{D}_{i,j}}}}$as an example:
\begin{equation}
\frac{{\partial L}}{{\partial {h_{C{D_{ij}}}}}} = \frac{{\partial L}}{{\partial {p_{tj}}}} \cdot \frac{{\partial {p_{tj}}}}{{\partial {z_{ij}}}} \cdot \frac{{\partial {z_{ij}}}}{{\partial {h_{C{D_{ij}}}}}}
\end{equation}
and the same process is also suitable to the other parameter, given space limitations, no further details are given here. As the Fig~\ref{Fig.softmax}, since the softmax function contains the output of all previous layers, a partial derivative is obtained for all n layers.

The result of  first term is as follows:
\begin{equation}
\frac{{\partial L}}{{\partial {p_{tj}}}} =  - \sum\limits_{j = 1}^n {\sum\limits_{i,t = 1}^c {\frac{{{y_{ij}}}}{{{p_{tj}}}}} }
\end{equation}
The second term of deriving the softmax function $\frac{{\partial {p_{tj}}}}{{\partial {z_{ij}}}}$ is more involved and needs to consider two situations:

(1) Deriving the current node when$t=i$:
\begin{equation}
\begin{array}{l}
\frac{{\partial {p_{tj}}}}{{\partial {z_{ij}}}} = \frac{{\partial \left( {\frac{{{e^{{z_{ij}}}}}}{{\sum\nolimits_{t = 1}^c {{e^{{z_{tj}}}}} }}} \right)}}{{\partial {z_{ij}}}} = \frac{{{e^{{z_{ij}}}} \cdot \sum\nolimits_{t = 1}^c {{e^{{z_{tj}}}}}  - {e^{{z_{ij}}}} \cdot {e^{{z_{ij}}}}}}{{{{(\sum\nolimits_{t = 1}^c {{e^{{z_{tj}}}}} )}^2}}}\\
\quad \quad = {p_{ij}} \cdot (1 - {p_{ij}})
\end{array}
\end{equation}

\begin{figure}[htbp]
\centering
\includegraphics[scale=0.7]{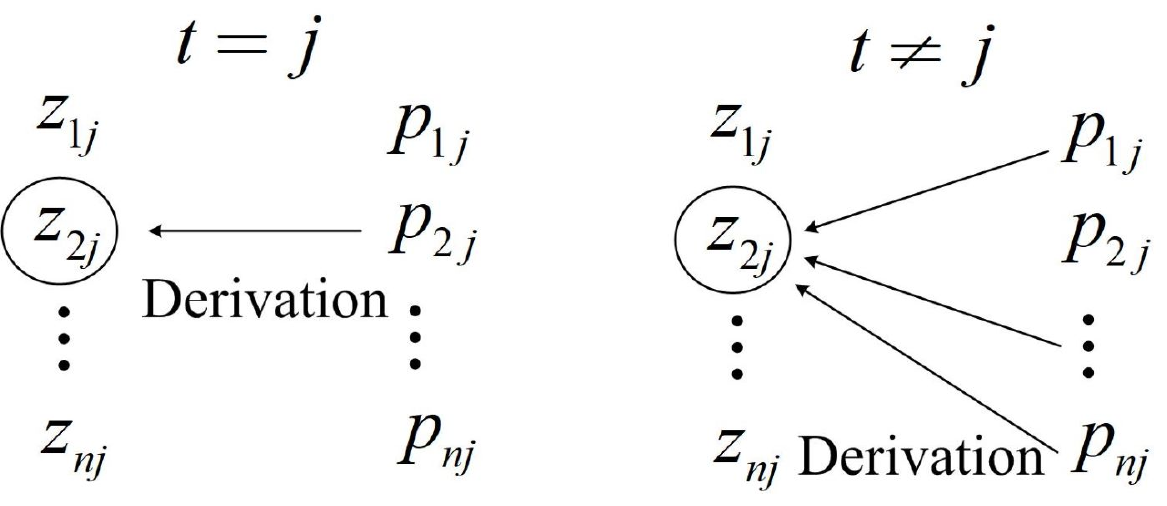}
\caption{Two situations for deriving ${{z}_{2j}}$}
\label{Fig.derivation}
\end{figure}

(2) Deriving the other nodes when$t \ne i$:
\begin{equation}
\frac{\partial {{p}_{tj}}}{\partial {{z}_{ij}}}=\frac{\partial \left( \frac{{{e}^{{{z}_{tj}}}}}{\sum\nolimits_{t=1}^{c}{{{e}^{{{z}_{tj}}}}}} \right)}{\partial {{z}_{ij}}}=-\frac{{{e}^{{{z}_{tj}}}}}{\sum\nolimits_{t=1}^{c}{{{e}^{{{z}_{tj}}}}}{{}^{2}}}\cdot {{e}^{{{z}_{ij}}}}=-{{p}_{tj}}\cdot {{p}_{ij}}
\end{equation}
We take an example for deriving ${{z}_{2j}}$ as the Fig~\ref{Fig.derivation} to explain the two situations.

And the third term is ${{b}_{C{{D}_{i,j}}}}$.
Then combine the results of the above three items, we have:
\begin{equation}
\begin{array}{l}
\frac{{\partial L}}{{\partial {h_{C{D_{ij}}}}}} =  - \sum\limits_{j = 1}^n {\sum\limits_{i,t = 1}^c {{y_{ij}} \cdot \frac{1}{{{p_{tj}}}} \cdot \frac{{\partial {p_{tj}}}}{{\partial {z_{ij}}}} \cdot {b_{C{D_{i,j}}}}} } \\
 = \sum\limits_{j = 1}^n {\sum\limits_{i,t = 1}^c {[ - \frac{{{y_{ij}}}}{{{p_{ij}}}} \cdot {p_{ij}} \cdot (1 - {p_{ij}}) + \sum\limits_{t \ne i} {\frac{{{y_{tj}}}}{{{p_{tj}}}} \cdot {p_{tj}} \cdot {p_{ij}}} ] \cdot {b_{C{D_{i,j}}}}} } \\
 = \sum\limits_{j = 1}^n {\sum\limits_{i,t = 1}^c {( - {y_{ij}} + {p_{ij}}\sum\limits_t {{y_{tj}}} ) \cdot {b_{C{D_{i,j}}}}} }
\end{array}
\end{equation}
Because $Y$ is a label matrix with 0 or 1 elements, And for every sample, $Y$satisfies the constraint$\sum\limits_{t}{{{y}_{tj}}}=1$.
So the partial derivative of L to ${{h}_{C{{D}_{i,j}}}}$can be written as follows:
\begin{equation}
\begin{array}{l}
\frac{{\partial L}}{{\partial {h_{C{D_{ij}}}}}} = \sum\limits_{j = 1}^n {\sum\limits_{i = 1}^c {({p_{ij}} - {y_{ij}}) \cdot {b_{C{D_{i,j}}}}} } \\
 = \sum\limits_{j = 1}^n {\sum\limits_{i = 1}^c {(\frac{{{e^{\sum\nolimits_{k = 1}^{{k_1}} {{b_{C{D_{i,k}}}}{\rm{\cdot}}{h_{C{D_{j,k}}}}}  + \sum\nolimits_{k = 1}^{{k_3}} {b_{S{D_{i,k}}}^{(v)}{\rm{\cdot}}h_{_{S{D_{j,k}}}}^{(v)}} }}}}{{\sum\nolimits_{t = 1}^c {{e^{\sum\nolimits_{k = 1}^{{k_1}} {{b_{C{D_{t,k}}}}{\rm{\cdot}}{h_{C{D_{j,k}}}}}  + \sum\nolimits_{k = 1}^{{k_3}} {b_{S{D_{t,k}}}^{(v)}{\rm{\cdot}}h_{_{S{D_{j,k}}}}^{(v)}} }}} }}} } \\
\quad \quad \quad \quad  - {y_{ij}}) \cdot {b_{C{D_{i,j}}}}
\end{array}
\end{equation}
Then the partial derivative of the column vector ${{h}_{C{{D}_{i}}}}$ is
\begin{equation}
\frac{{\partial L}}{{\partial {h_{C{D_i}}}}} = \sum\limits_{i = 1}^n {{{({p_i} - {y_i})}^T} \cdot {b_{C{D_i}}}}
\end{equation}
where ${{p}_{i}}$ is calculated by the softmax function.
Let ${{[\,]}_{+}}$ to denote $max\left( 0,\cdot  \right)$, which projects the negative value to the boundary of feasible region of zero and guarantees the non-negative nature of the matrix. And we get the update process for each matrix with learning rate $\eta $ as follows:
\begin{equation}
\begin{array}{l}
{w_{C{D_i}}} = {w_{C{D_i}}} + \eta [\sum\nolimits_{v = 1}^{{n_v}} {({R^{(v)}}{h_{C{D_i}}} - \alpha ({W_{CD}}{1_{k1 \times 1}}} \\
\quad \quad \quad \quad \quad \quad \quad \quad + W_{SD}^{(v)}{1_{k3 \times 1}}){]_ + }
\end{array}
\end{equation}
\begin{equation}
{w_{C{N_i}}} = {w_{C{N_i}}} + \eta {\left[ {\sum\nolimits_{v = 1}^{{n_v}} {{R^{(v)}}{h_{C{N_i}}}} } \right]_ + }
\end{equation}
\begin{equation}
w_{S{D_i}}^{(v)} = w_{S{D_i}}^{(v)} + \eta {\left[ {{R^{(v)}}h_{S{D_i}}^{(v)} - \alpha ({W_{CD}}{1_{k1 \times 1}} + W_{SD}^{(v)}{1_{k3 \times 1}})} \right]_ + }
\end{equation}
\begin{equation}
w_{S{N_i}}^{(v)} = w_{S{N_i}}^{(v)} + \eta {\left[ {{R^{(v)}}h_{S{N_i}}^{(v)}} \right]_ + }
\end{equation}
\begin{equation}
\begin{array}{l}
{h_{C{D_i}}} = {h_{C{D_i}}} + \eta [\sum\nolimits_{v = 1}^{{n_v}} {({R^{(v)}}{w_{C{D_i}}} - \frac{\beta }{{\rm{2}}}{1_{n \times 1}})} \\
\quad \quad \quad \quad \quad \quad \quad  - \frac{\gamma }{{\rm{2}}}{Q^{(v)}}^T{b_{C{D_i}}}){]_ + }
\end{array}
\end{equation}
\begin{equation}
{h_{C{N_i}}} = {h_{C{N_i}}} + \eta {\left[ {\sum\nolimits_v^{{n_v}} {{R^{(v)T}}{w_{C{N_i}}}} } \right]_ + }
\end{equation}
\begin{equation}
h_{S{D_i}}^{(v)} = h_{S{D_i}}^{(v)} + \eta {\left[ {{R^{(v)T}}w_{S{D_i}}^{(v)} - \frac{\beta }{2}{1_{n \times 1}} - \frac{\gamma }{2}{Q^{(v)T}}b_{SD}^{(v)}} \right]_ + }
\end{equation}
\begin{equation}
h_{S{N_i}}^{(v)} = h_{S{N_i}}^{(v)} + \eta {\left[ {{R^{(v)T}}w_{S{N_i}}^{(v)}} \right]_ + }
\end{equation}
Where ${{R}^{(v)}}$ and ${{Q}^{(v)}}$ are as follows:
\begin{equation}
{R^{(v)}}{\rm{ = }}{X^{(v)}} - {W_{CD}}H_{CD}^T - {W_{CD}}H_{CD}^T - W_{SD}^{(v)}H_{SD}^{(v)T} - W_{SN}^{(v)}H_{SN}^{(v)T}
\end{equation}
\begin{equation}
{Q^{(v)}} = {\rm{soft}}\max ({B_{CD}}H_{CD}^T + B_{SD}^{(v)}H_{SD}^{(v)T}) - Y
\end{equation}
Where $\eta $ denotes the learning rate, and the second derivative of the objective function can be used in the actual calculation process, which is closest to the best step size.
Furthermore, when the other variables are fixed, the projection matrix ${{B}_{CD}}$ and $B_{SD}^{(v)}$ could be calculated. Let the cross-entropy loss function be zero, and we can get the projection matrix ${{B}_{CD}}$ and $B_{SD}^{(v)}$ as follows:
\begin{equation}
{B_{CD}} = \frac{1}{{{n_v}}}\sum\nolimits_{v = 1}^{{n_v}} {\left( {Y - B_{SD}^{(v)}H_{SD}^{(v)T}} \right){H_{CD}}} {\left( {H_{CD}^T{H_{CD}} + \lambda I} \right)^{ - 1}}
\end{equation}
\begin{equation}
B_{SD}^{(v)} = \left( {Y - {B_{CD}}H_{CD}^T} \right)H_{SD}^{(v)}{\left( {H_{SD}^{(v)T}H_{SD}^{(v)} + \lambda I} \right)^{ - 1}}
\end{equation}
Where $I$is the identity matrix and $\lambda $ is a small positive number.
The specific processes of the block coordinate descent are depicted in Algorithm ~\ref{alg:A}.

\begin{algorithm}
\caption{MV-DLCSL}
\label{alg:A}
\begin{algorithmic}[1] 
\REQUIRE ~~\\ 
Multi-view data matrix ${X^{(1)}},{X^{(2)}}, \ldots ,{X^{({n_v})}}$, label matrix $Y$, parameters $\alpha ,\beta ,\gamma $, number of latent factors $k1,k2,k3,k4$;\\
\ENSURE ~~\\ 
Basis matrixes $W = \left\{ {{W_{CD}},{W_{CN}},W_{SD}^{(v)},W_{SN}^{(v)}} \right\}$;\\
Coefficient matrixes $H = \left\{ {{H_{CD}},{H_{CN}},H_{SD}^{(v)},H_{SN}^{(v)}} \right\}$;\\
Projection matrixes $B = \left\{ {{B_{CD}},B_{SD}^{(v)}} \right\}$;
\STATE Initialize matrixes $W,H$ and $B$;
\REPEAT
\STATE Update each column of ${W_{CD}}$ using Eq.(19);
\STATE Update each column of ${W_{CN}}$ using Eq.(20);
\FOR{$v = 1$ to ${n_v}$}
\STATE Update each column of $W_{SD}^{(v)}$ using Eq.(21);
\ENDFOR
\FOR{$v = 1$ to ${n_v}$}
\STATE Update each column of $W_{SN}^{(v)}$ using Eq.(22);
\ENDFOR
\STATE Update each column of ${H_{CD}}$ using Eq.(23);
\STATE Update each column of ${H_{CN}}$ using Eq.(24);
\FOR{$v = 1$ to ${n_v}$}
\STATE Update each column of $H_{SD}^{(v)}$ using Eq.(25);
\ENDFOR
\FOR{$v = 1$ to ${n_v}$}
\STATE Update each column of $H_{SN}^{(v)}$ using Eq.(26);
\ENDFOR
\STATE Update ${B_{CD}}$ using Eq.(29);
\FOR{$v = 1$ to ${n_v}$}
\STATE Update $B_{SD}^{(v)}$ using Eq.(30);
\ENDFOR
\UNTIL{convergence or reach max iterations}
\end{algorithmic}
\end{algorithm}

\section{Experiment}
In this section, we experimentally evaluate the proposed MV-DLCSL algorithm in classification task on seven real-world multi-view data sets, and analyze the convergence of our proposed block coordinate descent algorithm.

\subsection{Datasets}
In this paper, we use seven real-world multi-view data sets to verify the performance of the proposed algorithm, including Reuters, YaleFace, BBC, Cornell, Texas, Washington, and Wisconsin datasets. And Cornell, Texas, Washington, and Wisconsin dataset are four subset of data sets selected from WebKB data sets. The properties of data sets are summarized in Table 1.

\begin{table*}[]
\caption{Characteristics of the datasets}
\begin{tabular}{c|c c c c}
\hline
\multirow{2}{*}{\textbf{Data Set}} & \multicolumn{4}{c}{\textbf{Characteristics}}                                                                                          \\ \cline{2-5}
                                   & \textbf{\begin{tabular}[c]{@{}c@{}}The numbers of \\ Instances\end{tabular}} & \textbf{Views} & \textbf{Classes} & \textbf{Dimensions} \\ \hline
Reuters                            & 1200                                                                         & 5              & 6                & 2000 for all        \\
YaleFace                           & 256                                                                          & 2              & 8                & 2016 for all        \\
BBC                                & 685                                                                          & 4              & 5                & 4659/4633/4665/4686 \\
Cornell                            & 195                                                                          & 2              & 5                & 1703/585            \\
Texas                              & 187                                                                          & 2              & 5                & 1703/561            \\
Washington                         & 230                                                                          & 2              & 5                & 1703/690            \\
Wisconsin                          & 265                                                                          & 2              & 8                & 2703/795            \\ \hline
\end{tabular}
\end{table*}

\begin{table*}[]
\caption{Accuracy of different methods}
\begin{tabular}{c|c c c c c c c}
\hline
\multirow{2}{*}{\textbf{Method}} & \multicolumn{7}{c}{\textbf{ACC(\%)}}                                                                                          \\ \cline{2-8}
                                   & \textbf{Reuters} & \textbf{YaleFace} & \textbf{BBC} & \textbf{Cornell} & \textbf{Texas} & \textbf{Washington}& \textbf{Wisconsin}\\ \hline
GNMF          & 40.8$\pm$1.2 &50.0$\pm$2.5   &38.0$\pm$1.5   &41.0$\pm$1.8     & 57.9$\pm$1.8 &69.6$\pm$2.2 &52.8$\pm$1.4        \\
MultiNMF      & 52.7$\pm$0.2 &64.2$\pm$4.2   &73.1$\pm$0.2   &49.7$\pm$7.7     & 68.7$\pm$3.4 &59.3$\pm$2.6 &50.3$\pm$3.5        \\
MVCC          & 54.4$\pm$1.9 &33.3$\pm$6.9   &\textbf{95.8$\pm$2.6}   &60.8$\pm$5.0     & 64.7$\pm$5.5 &62.8$\pm$3.8 &64.3$\pm$2.7        \\
DICS          & 70.3$\pm$4.0 &89.1$\pm$3.2   &90.2$\pm$2.4   &72.8$\pm$6.1     & 81.6$\pm$4.0 &77.4$\pm$6.0 &\textbf{85.1$\pm$4.5}       \\
MV-DLCSL      &\textbf{70.5$\pm$1.3} &\textbf{90.2$\pm$1.5}   &91.7$\pm$1.2   &\textbf{75.3$\pm$2.2}     &\textbf{83.8$\pm$2.1} &\textbf{83.5$\pm$2.2} &83.6$\pm$1.9       \\ \hline
\end{tabular}
\end{table*}

\begin{figure*}[htbp]
\centering
\subfigure[BBC]{
\includegraphics[width=4cm,height=3cm]{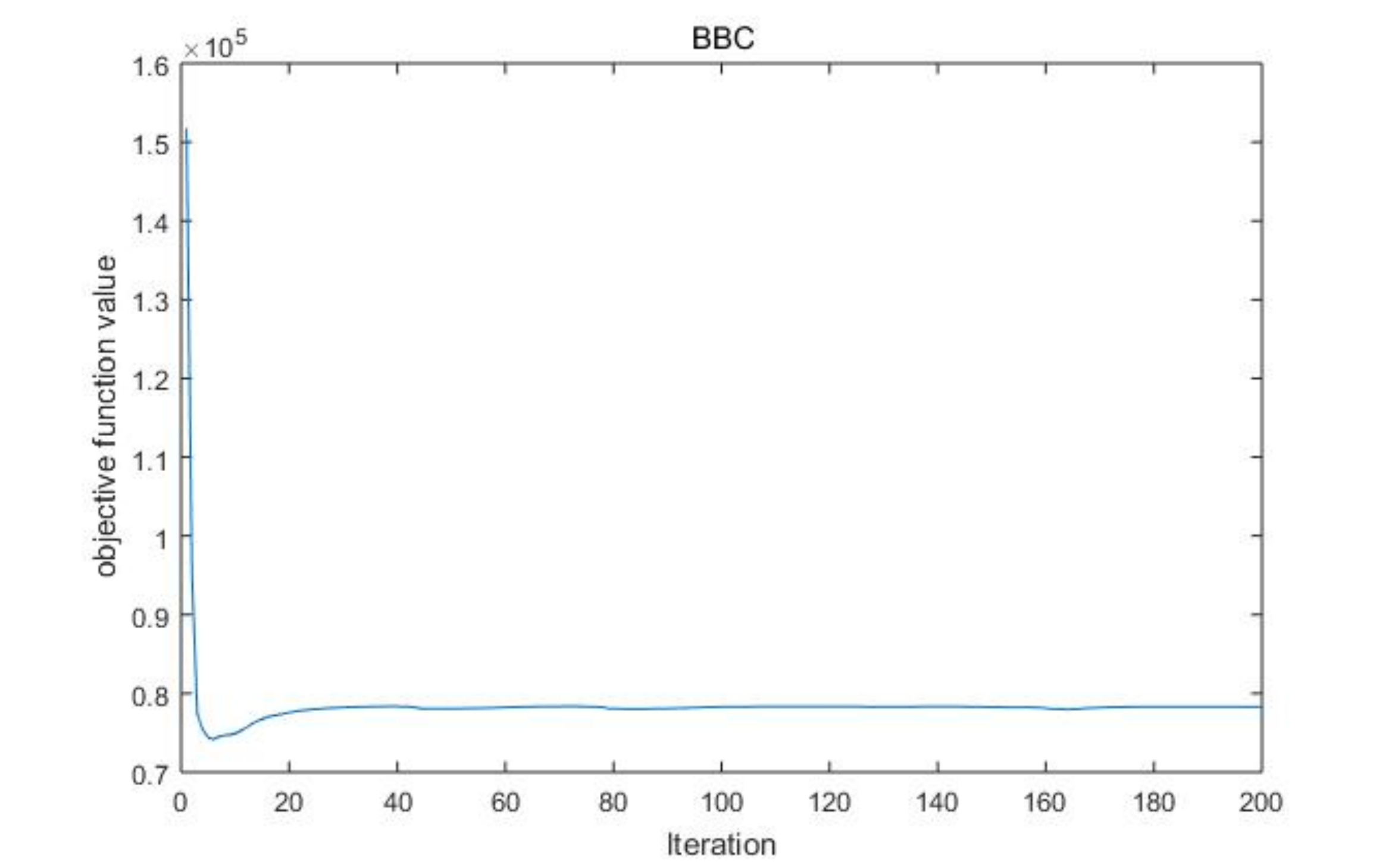}}
\quad
\subfigure[Reuters]{
\includegraphics[width=4cm,height=3cm]{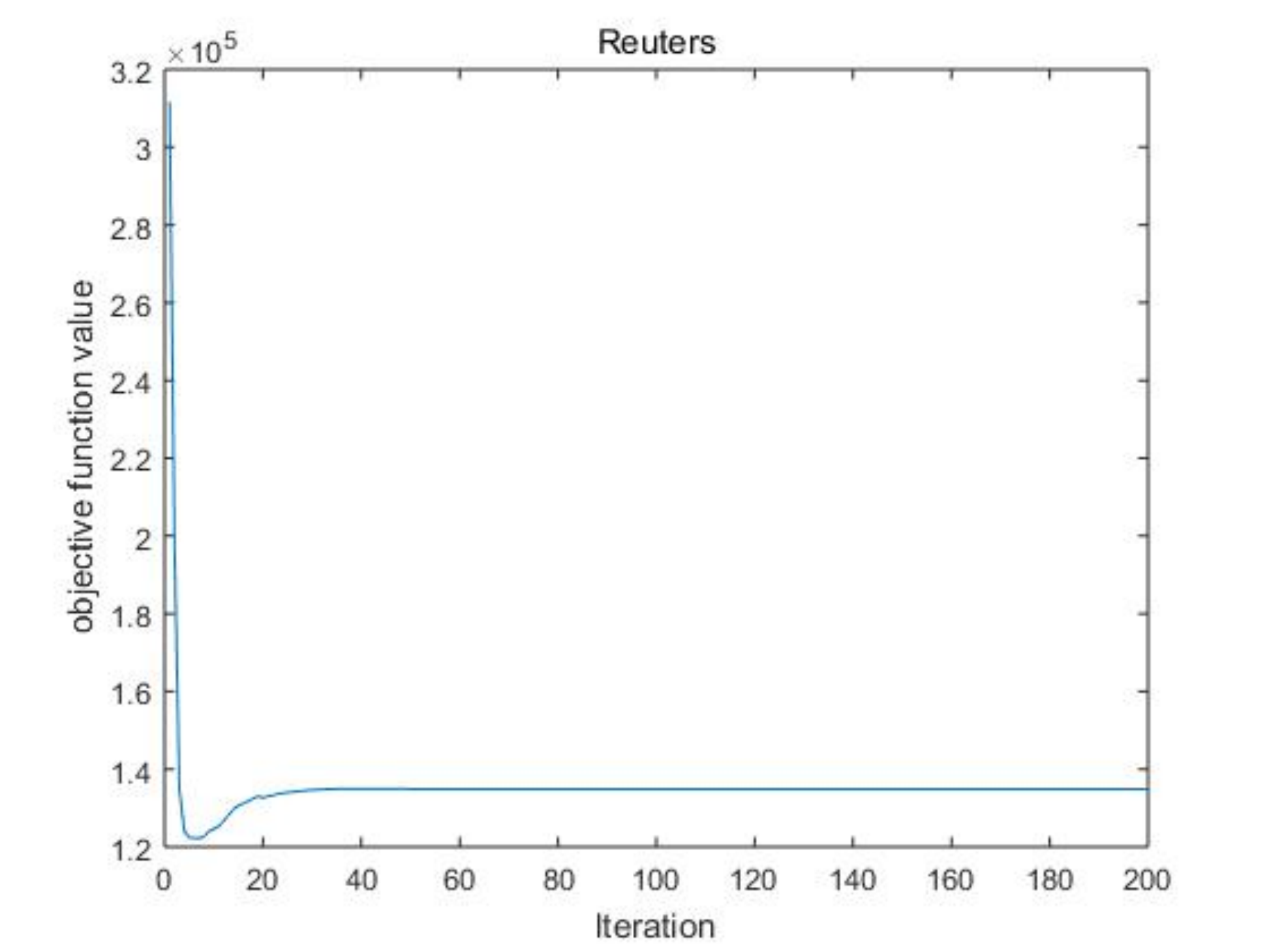}}
\quad
\subfigure[YaleFace ]{
\includegraphics[width=4cm,height=3cm]{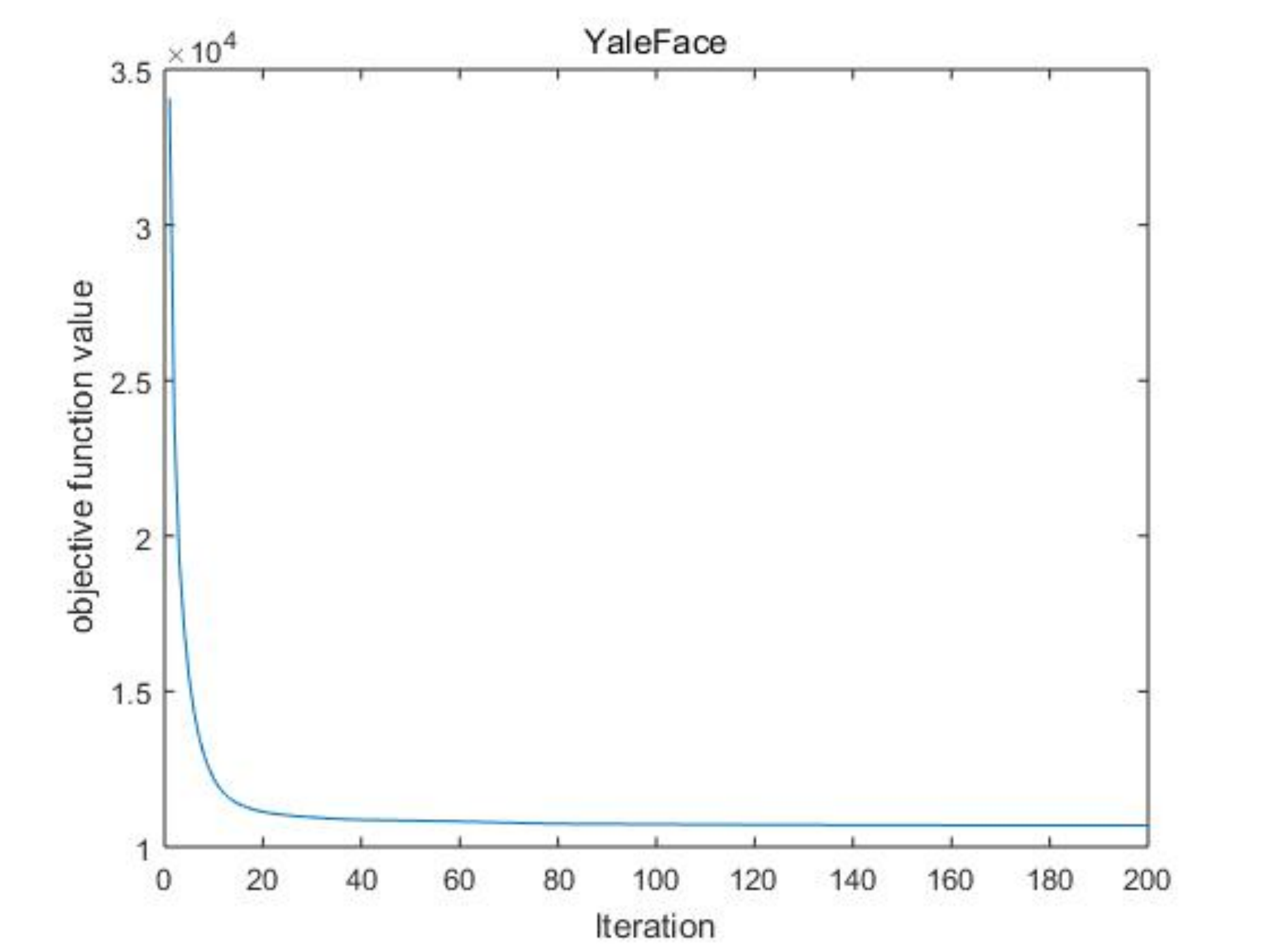}}
\caption{Iterative curves on all data sets}
\label{Fig.obj}
\end{figure*}
\subsection{Comparison Algorithms}
We compare our algorithm with several single-view and multi-view algorithms to show its effectiveness, including KNN, NMF, SSNMF, GNMF, multiNMF, MVCC, MCL, and DICS. And the parameters of all algorithms are selected within the range that the author suggested.

KNN(set $k=1$, i.e., 1-nearest neighbor classifier.) is regarded as the baseline algorithm and we apply KNN on  all single views and report the best performance on the view. Also we apply the KNN algorithm on the concatenated feature vector(KNNcat) which is extracted by baseline unsupervised comparison algorithms .

NMF is applied on each of the single view data and the concatenated feature vector (i.e. NMFcat) extracted by NMF on the single view data, which is regarded as inputs another baseline classification algorithm, such as1-nearest neighbor classifier.

SSNMF is a supervised NMF variant~\cite{35}, which incorporates a linear classifier to encode the supervised information. We select the regularization parameter $\lambda $ within the range of [0.5:0.5:3].

GNMF is a manifold regularized version of NMF~\cite{3}, which preserves the local similarity by imposing a graph Laplacian regularization. We use the normalized dot product (cosine similarity) to construct the affinity graph, and select the regularization parameter $\lambda $ within the set of $\{ {10^0},{10^1},{10^2},{10^3},$ ${10^4}\} $.

MultiNMF is an NMF-based multi-view clustering algorithm~\cite{11}, which can get compatible clustering results across multiple views. We select the regularization parameter $\lambda $ within the set of $\{10^0,10^{-1},$\\$10^{-2},10^{-3}\}$.

MVCC is a novel multi-view clustering method based on concept factorization with local manifold regularization~\cite{26}, which drives a common consensus representation for multiple views. We set parameter $\alpha $ to 100, and select $\beta $ and $\gamma $ within the set of $\{50,100,200,500,1000\}$.

MCL is a semi-supervised multi-view NMF variant with graph regularized constraint~\cite{4}. We select parameter $\alpha $ within the range of [100:50:250], $\beta $ within the set of ${\rm{\{ }}0.01,{\rm{ }}0.02,$ $0.03{\rm{\} }}$, and set $\gamma $ to 0.005 as author suggested.

DICS is the prototype model of our MV-DLCSL algorithm, which is an NMF-based multi-view learning algorithm, by exploring the discriminative and non-discriminative information existing in common and view-specific parts among different views via joint non-negative matrix factorization, and produce discriminative and non-discriminative feature from all subspaces. What's more, discriminative and non-discriminative features are further used to produce classification results. We select parameters $\alpha $ and $\beta $ within a small range of $\left[ 0,1 \right]$, and set parameter $\gamma $ to 1.

\subsection{Result}
For all algorithms, we first perform a five-folds cross validation to select the parame-ters that has the best accuracies and generalization performance. Due to randomness, we run all algorithms 10 times on each dataset and report the mean values and standard deviation of accuracies.
All the classification results of seven multi-view datasets are summarized in Table 2, and the best result on each dataset is highlighted in boldface. As we can see, the proposed algorithm achieves better accuracy on most of the datasets, and slightly worse than other algorithms on BBC and Wisconsin datasets, and It is worth mentioning that, the standard deviations of accuracies for  MV-DLCSL are much lower than DICS method.
\begin{figure*}[htbp]
\centering
\subfigure[DICS on BBC data sets (left panel) and our MV-DLCSL (right panel)]{
\includegraphics[width=0.45\textwidth]{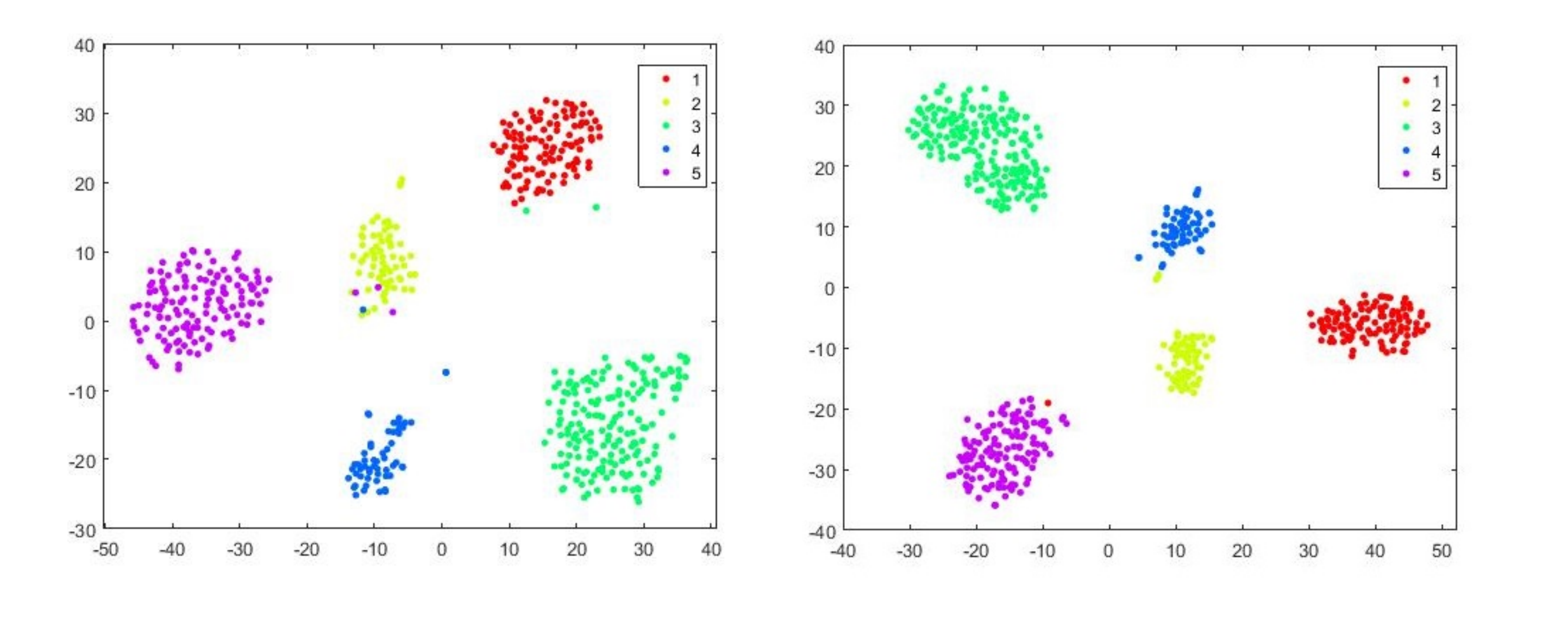}}
\quad
\subfigure[DICS on Cornell data sets (left panel) and our MV-DLCSL (right panel)]{
\includegraphics[width=0.45\textwidth]{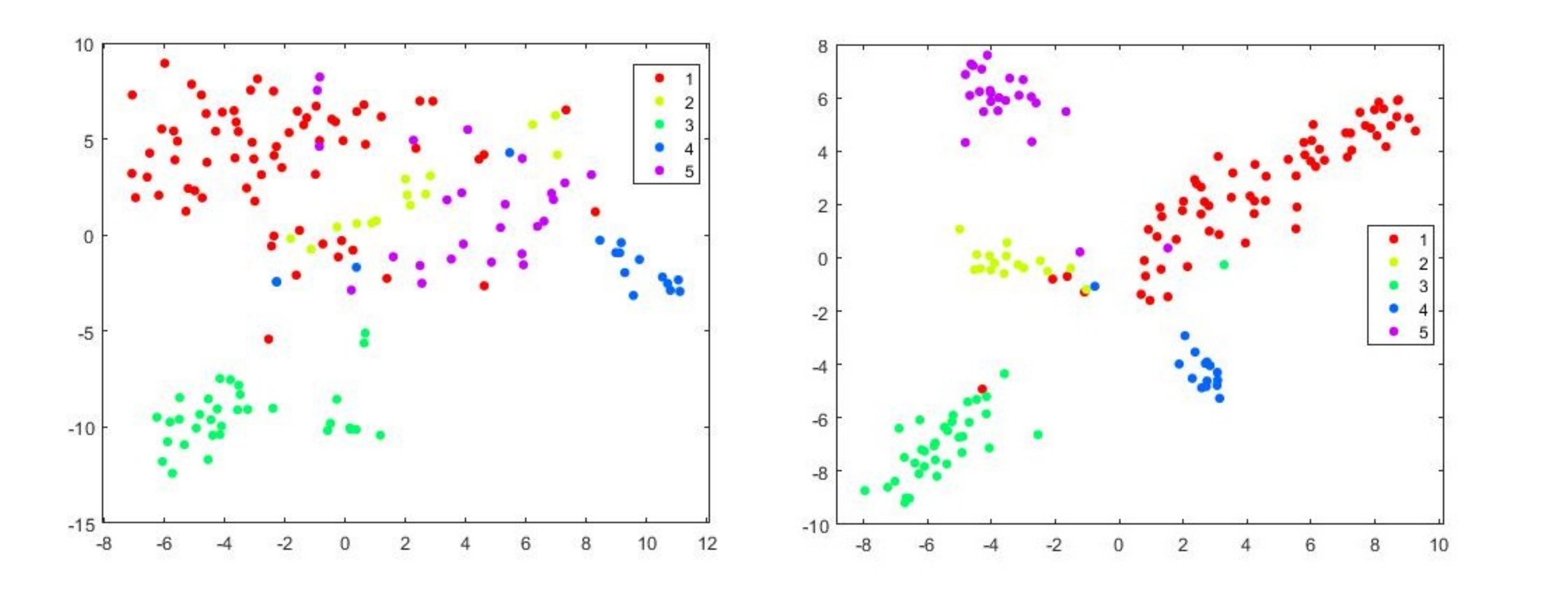}}
\quad
\subfigure[DICS on Reuters  data sets (left panel) and our MV-DLCSL (right panel)]{
\includegraphics[width=0.45\textwidth]{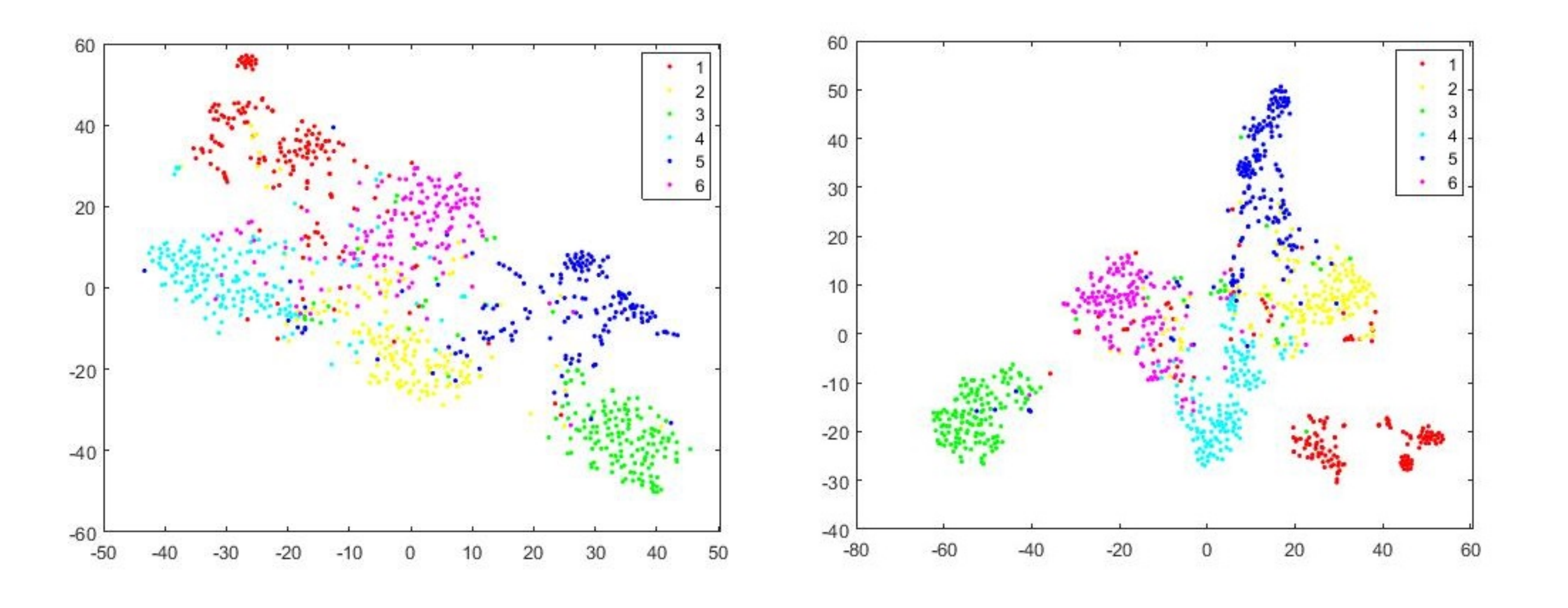}}
\quad
\subfigure[DICS on Texas data sets (left panel) and our MV-DLCSL (right panel)]{
\includegraphics[width=0.45\textwidth]{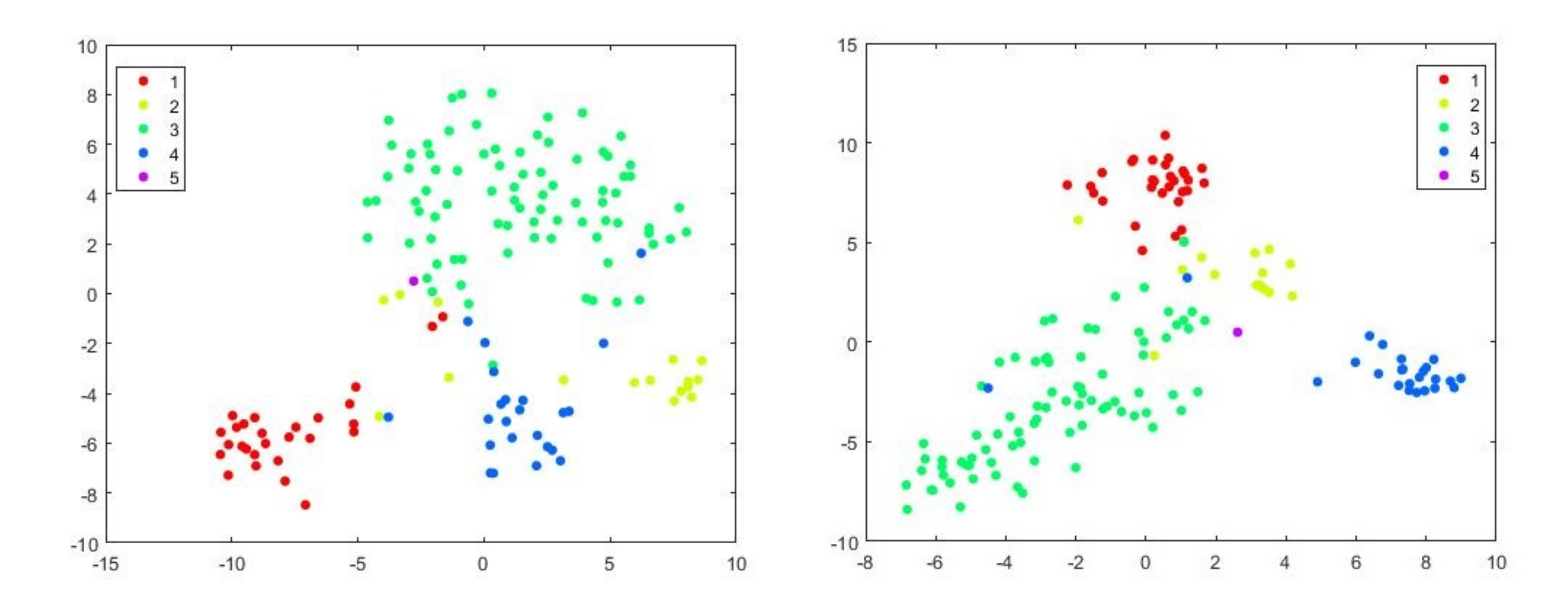}}
\quad
\subfigure[DICS on Washington data sets (left panel) and our MV-DLCSL (right panel)]{
\includegraphics[width=0.45\textwidth]{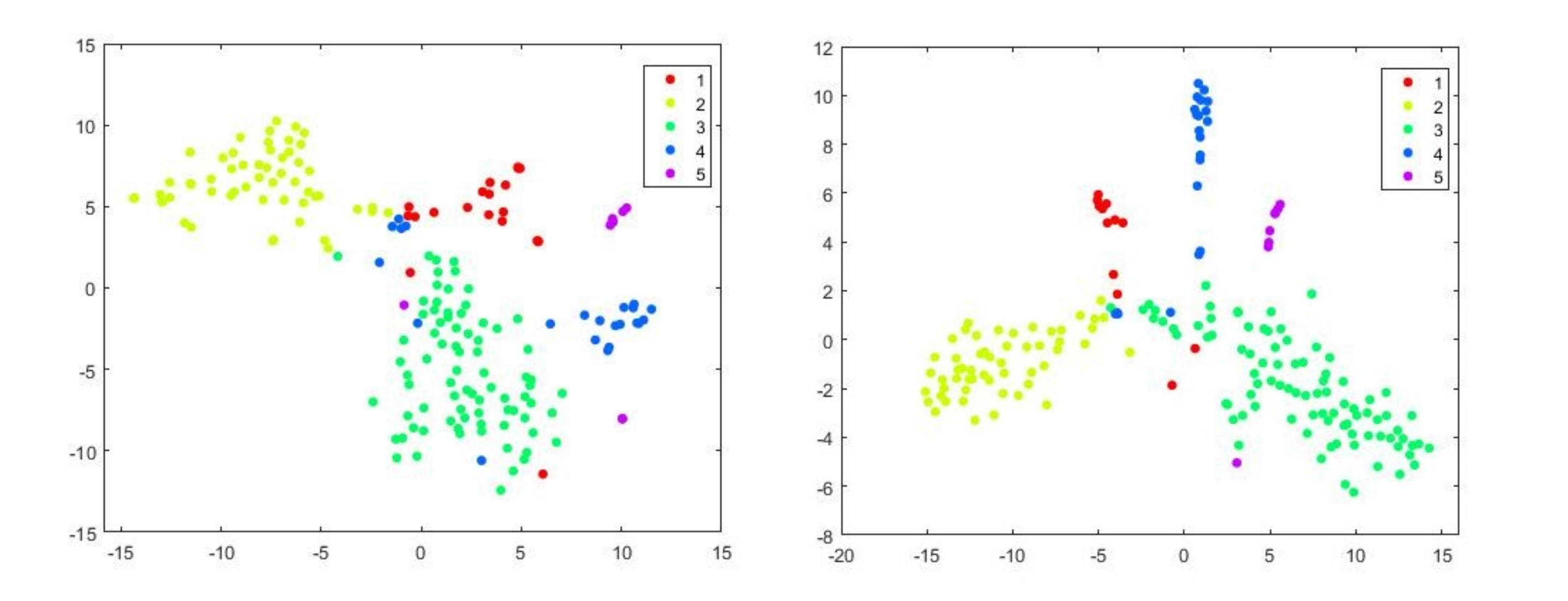}}
\quad
\subfigure[DICS on Wisconsin data sets (left panel) and our MV-DLCSL (right panel)]{
\includegraphics[width=0.45\textwidth]{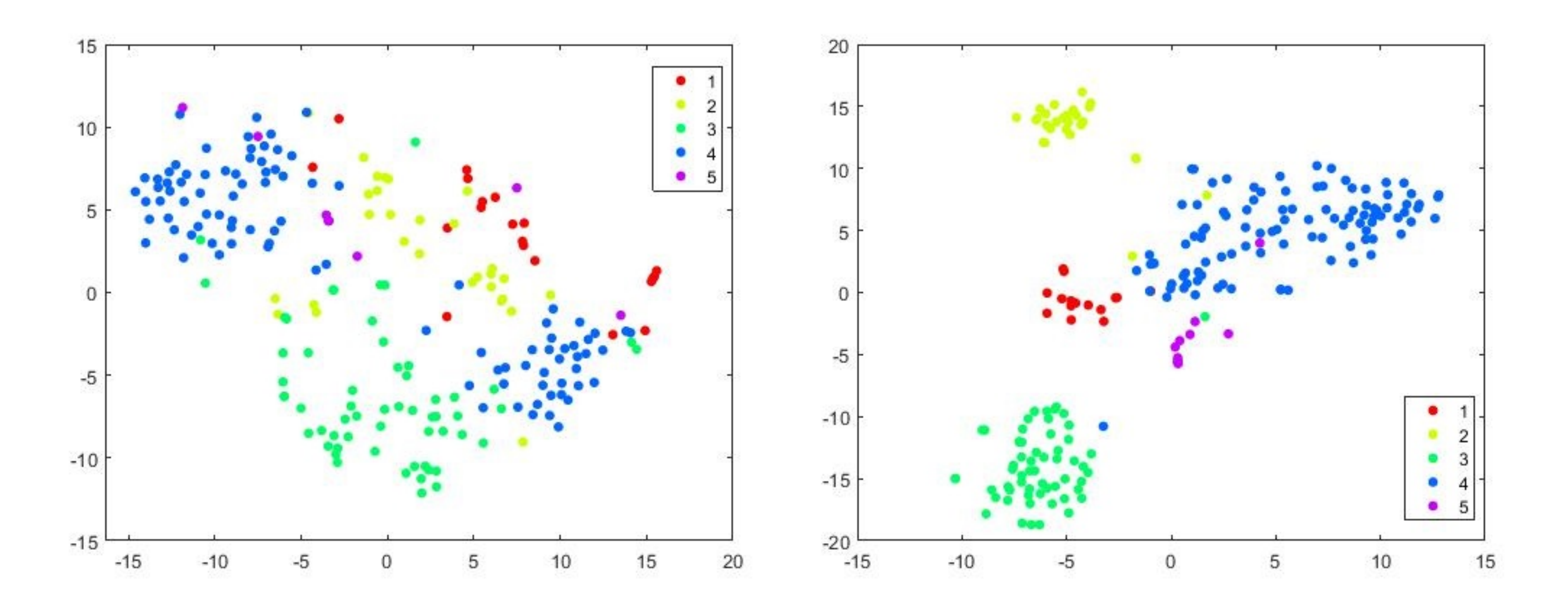}}
\quad
\subfigure[DICS on YaleFace data sets (left panel) and our MV-DLCSL (right panel)]{
\includegraphics[width=0.45\textwidth]{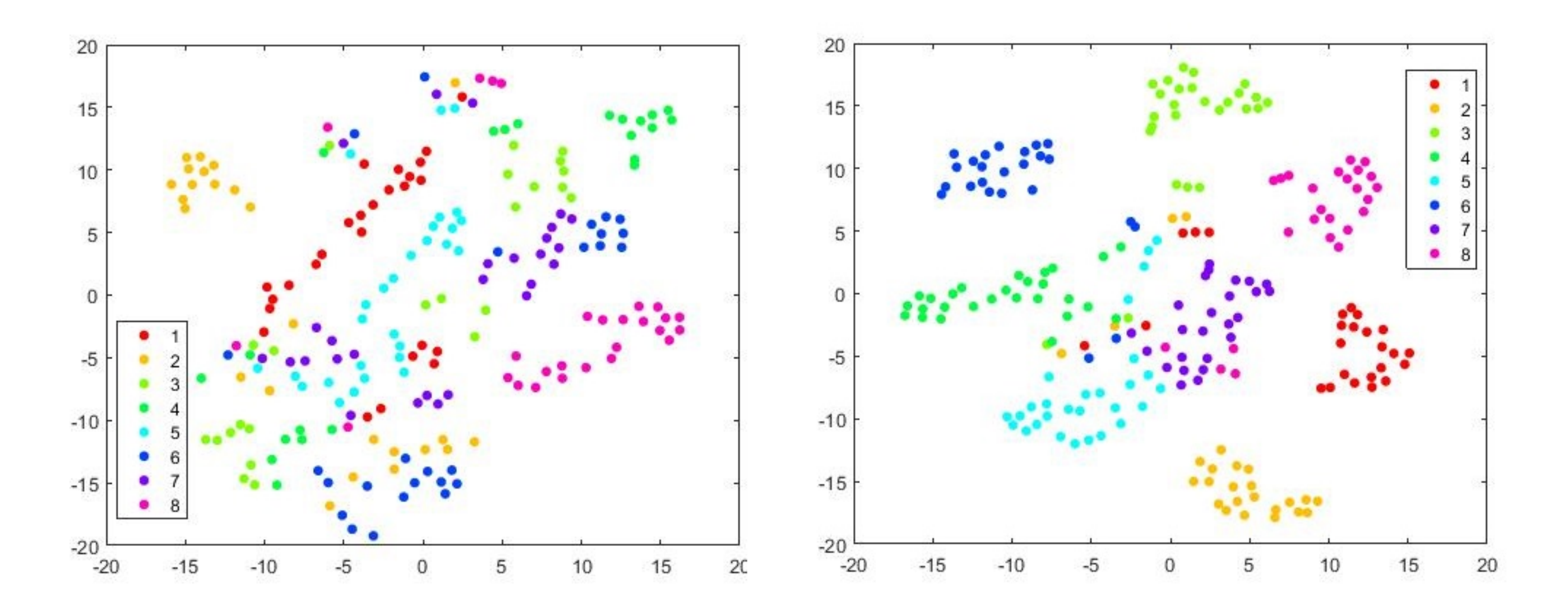}}

\caption{t-SNE scatter plot of the feature matrix on all data sets}
\label{Fig.tsne}
\end{figure*}

\subsection{Convergence Analysis}
In order to empirically investigate the convergence property of our algorithm, we plot the iterative curves of objective function on five typical data sets in Fig~\ref{Fig.obj}. From Fig~\ref{Fig.obj}, we can observe that the objective function values drop sharply and then iterative curves begin to grow/decrease mildly, then it converges eventually. Usually, the algo-rithm will converge in no more than 100 iterations.

\subsection{Discriminant Matrix Visualization}
t-SNE (t-distributed stochastic neighbour embedding) is a machine learning algorithm used for dimensionality reduction~\cite{36}. It was proposed by Laurens van der Maaten and Geoffrey Hinton in 2008. In addition, t-SNE is a nonlinear dimensionality reduction algorithm, which is very suitable for high-dimensional data dimensionality reduction to 2D or 3D for visualization.
We utilize DICS and our MV-DLCSL to obtain the basis matrix ${{W}_{D}}$ and the coefficient matrix ${{H}_{D}}$ through iterative updating processes, which is used to generate the feature matrix of raw data matrix$X$. By using the t-SNE for dimension reduction processing of the feature matrix, it is possible to visually see the distinguishable feature matrix. Therefore, we compare DICS with our MV-DLCSL method by two-dimensional visualized feature matrix to discern the classification capacities of feature matrices extracting by two methods. In the scatter plot generated by t-SNE, each point represents an instance and different colors indicate different class labels. Therefore, most points with the same color are clustered into clusters in the scatter plot. From Fig~\ref{Fig.tsne}, we could see the t-SNE scatter plot of feature matrices that our MV-DLCSL method extract having better clustering results, and there are fewer instances assigned to the wrong cluster. What's more, the clusters in our MV-DLCSL method are more concentrated than DICS. Thus, our method is able to find the feature matrix more accurately than DICS.

\section{Conclusions and Future Work}
In this paper, we propose a novel multi-view network MV-DLCSL. The proposed algorithm explores the discriminative and non-discriminative information existing in common and view-specific parts among different views via joint non-negative matrix factorization, and use the cross entropy loss to constrain the objective function, which shows better classification results than the mean square error. The experimental results on seven real-world data sets have demonstrated the effectiveness of our proposed algorithm.
For future studies, we plan to consider the distance farther between the view-specific matrixes to grasp the information related to the common and view-specific part, which is benefit to precisely classify instances through properties between different views.